%% file: main.tex
\documentclass[a4paper,11pt]{article}

\usepackage{graphicx}  
\usepackage{url}       
\usepackage{times}
\usepackage{natbib}
\usepackage[margin=25mm]{geometry}

\usepackage{amsmath}
\usepackage{todonotes} 
\usepackage{calc}	  
\usetikzlibrary{calc} 
\usepackage{tikz}     
\usepackage{tabularx}
\usepackage{arydshln} 
\usepackage{booktabs,multirow} 
\usepackage{array,ragged2e}    
\usepackage[normalem]{ulem} 
\usepackage[colorlinks=true,allcolors=blue!60!black,breaklinks]{hyperref} 

\usepackage{macrosDRS} 

\newcommand{\comments}[1]{} 

\title{Towards Universal Semantic Tagging}
\date{}

\author{Lasha Abzianidze\\
       CLCG, University of Groningen\\
       \texttt{l.abzianidze@rug.nl}
  \and Johan Bos\\
      CLCG, University of Groningen\\
       \texttt{johan.bos@rug.nl}
}

\begin{document}
\maketitle
\thispagestyle{empty}
\pagestyle{empty}

\begin{abstract}
The paper proposes the task of  universal semantic tagging---tagging word tokens with language-neutral, semantically informative tags.
We argue that the task, with its independent nature, contributes to better semantic analysis for wide-coverage multilingual text.   
We present the initial version of the semantic tagset and show that (a) the tags provide semantically fine-grained information, and (b) they are suitable for cross-lingual semantic parsing. 
An application of the semantic tagging in the Parallel Meaning Bank supports both of these points as the tags contribute to formal lexical semantics and their cross-lingual projection.
As a part of the application, we annotate a small corpus with the semantic tags and present new baseline result for universal semantic tagging.
\end{abstract}

\section{Introduction}

Part-of-speech (\pos{}) tagging represents one of the most popular Natural Language Processing (NLP) tasks, especially when it comes to syntactic parsing.
It is proven by practice that the information about \pos{}-tags makes syntactic parsing easier.
An independent nature of the task and its lower complexity (compared to syntactic parsing) make \pos{} tagging a perfect preprocessor for syntactic parsing.

But to what extent is \pos{}-tag information useful for semantic parsing---obtaining semantic representations of natural language texts?
Trying to answer this question in favor of \pos{}-tags, we take a stand of a semantic parsing approach that heavily relies on them.
One of such approaches is the formal compositional semantics driven by syntactic derivations of Combinatory Categorial Grammar (CCG, \citealt{Steedman:01}), 
where a meaning representation is derived by composing formal meaning representations of lexical items \citep{Bosetal:04,lewis-steedman:13,mineshima-EtAl:2015:EMNLP}.%
\footnote{Similarly, the semantic parsing based on dependency structures \citep{reddy_transforming_2016,reddy_USP_2017} also rely heavily on POS-tags. 
}
Since lexical items come with fully fledged semantics, obviously, assigning correct lexical semantics is crucial for this approach.
This is the place where \pos{}-tags come into play by providing lexical information helping to determine lexical semantics.
For example, given a \pos{}-tag \postag{NN} (singular or mass noun) or \postag{JJ} (adjective)%
\footnote{Throughout the paper the Penn Treebank \pos{}-tags \citep{Marcus:1993}, widely accepted in the NLP community, will be assumed unless otherwise stated. 
},
it is possible to assign the desired lexical semantics to the modifiers in \eqref{eq:NN_JJ}.
But there are cases where \pos{}-tags fall short of providing sufficient information for lexical semantics.
For instance, regardless of their semantics, quantifiers get the same tag \postag{DT} (determiner).
Hence one needs to check the lemma of a determiner in order to define its semantics in \eqref{eq:no_every}.

\smallskip
\noindent\begin{minipage}{.505\linewidth}
\begin{equation}
\kern-3mm
\raisebox{2mm}{\scalebox{.83}{
\CCG{
$N/N$\\
$\lam{px} beer(x) \wedge p(y) \wedge for(y,x)$ if pos=NN\\
$\lam{px} transparent(x) \wedge p(x)$ if pos=JJ 
}{beer$^\postag{NN}$ / transparent$^\postag{JJ}$}
\,
\CCG{
$N$\\
$\lam{x}bottle(x)$}
{bottle$^\postag{NN}$}
}}
\kern-2mm
\label{eq:NN_JJ}
\end{equation}
\end{minipage}%
\begin{minipage}{.50\linewidth}
\begin{equation}
\raisebox{2mm}{\scalebox{.83}{
\CCG{
$\NP/N$\\
if lemma=`no' $\lam{pq} \neg \exists x \big(p(x) \wedge q(x)\big)$\\
if lemma=`every' $\lam{pq} \forall x \big(p(x) \to q(x)\big)$
}{no$^\postag{DT}$ / every$^\postag{DT}$}
\CCG{
$N$\\
$\lam{x}man(x)$}
{man$^\postag{NN}$}
}}
\kern-3mm
\label{eq:no_every}
\end{equation}
\end{minipage}\bigskip

Formal semantics of a content word usually involves a symbol corresponding the lemma.
This is the case for each lexical item in \eqref{eq:NN_JJ} and \eqref{eq:no_every} except for the quantifiers.
But when dependence of lexical semantics on a lemma is beyond a simple substitution, i.e., one needs to verify a lemma to define lexical semantics, then this case fails to generalize across different languages.
For example, assigning lexical semantics to quantifiers based on their lemma does not scale up for multilingual semantics.
On the other hand, the treatment of common nouns in \eqref{eq:NN_JJ} and \eqref{eq:no_every} generalizes for a multilingual case by using a simple assignment: 
\begin{equation}
\sem{\langle w,~  \text{pos}=\postag{NN},~ \text{category} = N \rangle} = \lam{x} \textsc{sym}(x)
\label{eq:sem_assign}
\end{equation}
where \textsc{sym} is a lexical predicate, usually a lemma, corresponding to the word $w$.
%

In order to compensate the shortcomings of \pos{} tagging for semantic parsing, we propose a new NLP task, called  {\em Universal Semantic Tagging} or  {\em Semantic Tagging} in short.%
\footnote{Since semantics of linguistic expressions is language independent to a large extent, we find {\em universal} redundant from a semantic perspective.
On the other hand, from an NLP perspective, we would like to emphasize the universal (i.e., cross-lingual) nature of the task. 
}
The task represents a standard sequence tagging problem where each word token gets a language-neutral semantic tag, in short {\em sem-tag}.
Sem-tags carry information that better characterizes lexical semantics than \pos{}-tags do.   
We will show that the semantic tagging not only improves over \pos{} tagging but also subsumes the task of Named Entity (\namen{}) classification.  
We argue that importance of the task for (cross-lingual) semantic parsing is comparable to the one \pos{} tagging has for syntactic parsing.

The rest of the paper is organized as follows.
First, we further motivate the idea behind semantic tagging---how it includes semantic virtues of \pos{}-tags and Named Entity (\namen{}) classes;
Then we present the current version of the semantic tagset.
To show application of semantic tagging in semantic parsing, we describe its use in the Parallel Meaning Bank (PMB) project%
\footnote{\url{http://pmb.let.rug.nl}}
\citep{PMBshort:2017}, where the sem-tags help to determine formal lexical semantics.
We also present a baseline result for semantic tagging on a small annotated corpus.
In the end, the paper discusses possible directions of future research on semantic tagging.

\section{Motivation for Semantic Tagging}

The information about \pos{}-tags and \namen{} classes do contribute to determine lexical semantics to some extent, but they are not sufficiently informative.
One of the goals of the semantic tagging is to incorporate semantic virtues of these two tasks and fill gaps in semantic modeling by adding new categories. 

In a tagging task, a sequence of characters that takes a tag is called a {\em word token}, or simply a {\em token}.
Definition of a token may vary depending on a tagging task and its application.
We find the concepts of token for POS tagging and semantic tagging somewhat different.
For example, {\em ``20-year-old man from New Zealand''} represents five tokens for POS tagging while we consider six token version {\em ``20 year old man from New\tokglue{}Zealand''} more suitable for semantic analysis.%
\footnote{In general, we assume a fixed multiword expression as one token if it is semantically non-compositional and has an obscure syntactic structure.
Such multiword tokens include proper names (e.g., {\em Alfred\tokglue Nobel} and {\em European\tokglue Union}), numerical expressions (e.g., {\em ten\tokglue thousand} and {\em 10\tokglue 000}), and function phrases like {\em as\tokglue well\tokglue{} as}, {\em each\tokglue other}, and {\em so\tokglue that}.
}
Hereafter, when talking about semantic tagging, tokens should be understood as meaningful atoms.

%

In addition to the examples from the previous section, \pos{}-tags fail to disambiguate lexical semantics of series of word tokens.
For example, reflexive and emphasizing pronouns get the same \pos{}-tag \postag{PRP}.
The conjunctions {\em and}, {\em or} and {\em but} are all \pos{} tagged as coordinating conjunctions (\postag{CC}).%
\footnote{Moreover, there are at least two semantic usages of {\em and} one might want to distinguish: distributive and collective readings.
} 
A comma can have several semantic functions, e.g., \cite{comma-sem:16} distinguishes nine semantic roles including apposition, location or listing.
Both infinitival and prepositional uses of {\em to} are \pos{} tagged as \postag{TO} \citep[p.\,5]{POStagging:90}.
Semantics of the determiner {\em any} needs to be disambiguated in context.
The auxiliary verbs (e.g., {\em do} and {\em have}) and content verbs obtain similar \pos{}-tags based on their syntactic features.
This complicates to determine whether a verb introduce an event entity or not.
The relative pronouns {\em which} and {\em that} both get the \postag{WDT} \pos{}-tag regardless of their restrictive or non-restrictive behaviour.
It is natural to distinguish semantics of intersective adjectives (e.g., {\em ill} and {\em dead}) from subsective ones (e.g., {\em skillful} and {\em professional}), but this is impossible to do with the single \pos{}-tag \postag{JJ}.
The above-mentioned partial list clearly shows that \pos{}-tags are not sufficient for fine-grained (formal) lexical semantics.

For wide-coverage semantic analysis one needs to identify \namen{}s, detect their type, and  model their semantics appropriately.
The information extraction community has been actively working on the problem of \namen{} classification and designed annotation schemas.
For example, the named entity task at MUC-7 \citep{MUC7def:98} distinguished three general classes of \namen{}s, where each of them contain several types: entity names (person, organization, location), temporal expressions (date and time) and number expressions (money and percentage). 
These types of \namen{}s are motivated by downstream applications of information extraction.
For a fine-grained semantic analyses, one might go beyond this standards.
For example, one of such moves, following to \cite{ACE:04}, is to distinguish the locations without political or social groups (e.g., seas, parks and mountains) from those with them, i.e. geo-political entities such as villages, cities, countries, etc.
Also one can introduce new \namen{} classes, for instance, the classes for events (e.g., {\em 9/11} and {\em World War II}) and artifacts (e.g., {\em Ubuntu 12.04 LTS}) or generalize existing ones, for example, go beyond monetary currency and percentage and cover the measure words like {\em meter} and {\em kilogram}.

In the next section we present an inventory of the universal semantic tags which incorporates semantic merits of \pos{}-tags and named entity classes, fill the gaps in semantic annotation, and represents one unified tagset aiming to facilitate cross-lingual semantic parsing.

\section{The Universal Semantic Tagset}

The universal semantic tagset aims to provide general cross-lingual description for lexical semantics of all sorts of word tokens.
It significantly differs from \pos{} tagset, which is not semantically motivated, and generalizes over \namen{} classes as the latter only covers the words of a particular type.
The current version of the semantic tagset (v0.7) is given in \autoref{tab:tagset}, a revised version of the tagset (v0.6) presented in \cite{Bjervaetal:16}.%
\footnote{Major revisions concern the classes of named entity (\mttg{NAM}), attributes (\mttg{ATT}), events (\mttg{EVE}), deixis (\mttg{DXS}) and tense (\mttg{TNS}). 
In contrast to the tagset v0.6, the current tagset excludes 15 tags and includes 13 new ones.
More details about the changes are explained below.
}
The sem-tags are organized into 13 coarse-grained semantic classes each having its own meta-tag.
This division is informal as many sem-tags easily qualify for several classes.
%
We designed the tagset in a data-driven fashion while bearing in mind formal semantic properties of tokens.
The employed corpus consists of several parallel corpora of various genres spanning over four languages (see \autoref{sec:pmb_app}).

Before we characterize the sem-tags, let us explain how the tags can or cannot be interpreted.
A~sem-tag of a token describes a semantic contribution of the token with respect to the meaning of the source expression.
In this way, the principle of semantic compositionality underlies the semantic tagging.
Later, in \autoref{sec:pmb_app}, an application shows how to interpret a sem-tag as an unspecified semantic schema/recipe.
In general, sem-tags are not responsible for encoding a syntactic function of a token;
For example, concrete quantities and colors get \smtg{QUC} and \smtg{COL} regardless of being a nominal modifier or a head of a noun phrase.%
\footnote{In contrast to this, depending on a syntactic context a color can get the \postag{NN} or \postag{JJ} \pos{}-tag in the Penn Treebank \citep[p.\,12]{POStagging:90}: {\em The plants are dark green/}\postag{JJ} vs {\em The plants are a dark green/}\postag{NN}.
It is also needless to say that sem-tags do not distinguish singular or non-3rd person verb forms, unlike the \pos{}-tags. 
}
Moreover, currently sem-tags do not separate adjectives and adverbs but treat them as properties.
The information about thematic roles are not also provided by the sem-tags.
In principle, sem-tags provide the semantic information that complements thematic roles, syntax and lemma.
Due to the abstraction from syntactic and lemma-related information, sem-tags are suitable for cross-lingual application.

\begin{table}[!t]
\caption{The Universal Semantic Tagset v0.7: 73 sem-tags grouped into 13 meta-tags. 
The sem-tags are accompanied with the examples where several highly ambiguous tokens come with a \context{context}. The new sem-tags of v0.7 wrt v0.6 are marked with an asterisk.}
\scalebox{1}{
\footnotesize
\hspace{-21mm}
\input{tagset2}
}
\label{tab:tagset}
\end{table}

The semantic classes \mttg{ATT}, \mttg{COM}, \mttg{NAM}, \mttg{EVE} and \mttg{UNE} cover both open and closed class words while the rest of the classes focus on the closed class words.
The sem-tags that model closed class words make two major contributions: (i) semantically disambiguate highly ambiguous words that usually belong to closed class words, and (ii) act as an umbrella term for cross-lingual variants of a word and opens the door to multilingual semantic tools.         

Let us first describe the groups of sem-tags covering closed class words.
The anaphoric tags encompass definite articles and types of pronouns.
They distinguish emphasizing pronouns (\smtg{EMP}) from reflexive ones (\smtg{REF}).
Other types of determiners like indefinite articles, demonstratives, and quantifiers (most of which get the \postag{DT} pos-tag) are covered by existential (\smtg{DIS}), universal (\smtg{AND}), place deixis (\smtg{DXP}) and vague quantity (\smtg{QUV}) sem-tags. 
Besides place deixis, there are sem-tags for temporal and discourse deixis \citep[Ch.\,4]{lobner2013}.
In addition to the sem-tags for subordinated (\smtg{SUB}) and coordinated (\smtg{COO}) discourse relations, there are separate tags \smtg{APP} and \smtg{BUT} for appositional and contrasting relations.
Phrasal conjunctions and other discourse relations that have relatively transparent formal logical semantics are singled out by the logical sem-tags \smtg{DIS}, \smtg{IMP}, and \smtg{AND}.
Tokens with vacuous semantics are tagged with \smtg{NIL}.
Such tokens might include punctuations, infinitival {\em to}, and {\em of} from pseudo-partitives, e.g., {\em two liters of water}.
The \mttg{LOG} class also includes the tags \smtg{ALT} and \smtg{XCL} covering words with semantics involving inequality. 
Given these sem-tags, a comma might be tagged with \smtg{NIL}, \smtg{APP}, \smtg{AND}, or \smtg{DIS} depending on its semantic contribution.
Relative pronouns of restrictive and non-restrictive relative clauses get \smtg{AND} and \smtg{APP} respectively.

Accounting for modal words in semantics is crucial as they often block certain entailments.
For this reason, the tagset has dedicated tags for tokens with modal functions, including a tag \smtg{NOT} for negative lexical items.
In contrast, the Penn Treebank \pos{}-tagset distributes most of negative items among adverbs, prepositions and determiners.
Unlike the \pos{}-tags, the sem-tags distinguish tense and aspect marking auxiliary verbs (\mttg{TNS}) from content (\mttg{EVE}) and modal (\mttg{MOD}) ones.

Since date and time expressions play an important role is downstream applications and have been a target of several shared tasks, the tagset has fine-grained sem-tags for them: \smtg{DAT} and \smtg{CLO} are designed for the full date and time formats while the rest marks (unspecified) components of the date format.
We also design special sem-tags for speech acts.

The attributive and comparative classes mostly cover words like adjectives, adverbs, quantities and words derived from them.
Since both adjectives and adverbs can be seen as modifiers (of entities and events) from a semantic perspective, sem-tags do not differentiate them.%
\footnote{Moreover, some languages like Dutch and German make little grammatical distinction between adverbs and adjectives. 
}
The sem-tags distinguish concrete \smtg{QUC} and vague \smtg{QUV} quantities (which were previously merged in a single quantity sem-tag in v0.6).
There are separate tags for intersective, subsective and privative adjectives.
The adjective like {\em alleged} that are neither subsective nor privative are tagged with the modal sem-tag \smtg{POS}. 
Adverbs are usually tagged as intersective.
\smtg{DEG} marks adjectives that subcategorize for degrees
, e.g., {\em 10cm long} or {\em 2 years old}, 
as they are not subsective.
From comparative sem-tags, we would like to mention \smtg{EQU} which covers words with interpretation of (tense-free) equality.

The tagset makes fine-grained distinction of proper names.
In addition to the standard \namen{} tags \smtg{PER} and \smtg{ORG}, following \cite{ace2008events}, geographical locations are divided into geo-political entities (geographical regions defined by political and/or social groups, \smtg{GPE}) and the rest of geographical entities (\smtg{GEO}).
To link individuals to the \namen{}s they originate from, we use \smtg{GPO}.
Units in measure phrases are tagged with \smtg{UOM} as they act like \namen{}s.%
\footnote{It seems unnatural to treat them as predicates and therefore license entailments using the WordNet hypernymy relations: {\em he ran five kilometers} $\Rightarrow$ {\em he ran five metric linear units}. 
}
\smtg{UOM} generalizes over the standard \namen{} class for currency and percentage.

The event sem-tags account for semantics of content verbs that introduce Davidsonian event entities.
\smtg{EXS} marks a content verb without tense or aspect (including nominalizations and gerunds) while the other sem-tags in \mttg{EVE} additionally encode tense or aspect.
Currently, the sem-tags in \mttg{EVE} are motivated by English, German, Dutch and Italian.
The sem-tags for unnamed entities divide nouns into concepts (\smtg{CON}), roles (\smtg{ROL}), and collective/group nouns (\smtg{GRP}).
Moreover, \smtg{ROL} also covers relational nouns while \smtg{GRP} marks collective operators too.  

The examples of semantically tagged tokenized texts are given below.
In \eqref{sem-tag:1}, {\em tall} is marked with \smtg{DEG} as it is not affirmative---the question is not asking weather the green monster is tall.
The sem-tags in \eqref{sem-tag:3} and \eqref{sem-tag:4} disambiguate existential and universal semantics of {\em any} and {\em a}.
Notice that the latter is tagged with \smtg{AND} as {\em\$ 100 a day} is semantically equivalent to {\em\$ 100 each day}.
More examples of semantically tagged text can be accessed online via the PMB Explorer.%
\footnote{\url{http://pmb.let.rug.nl/explorer/}
}


\bgroup
\setlength\abovedisplayskip{2pt}
\setlength\belowdisplayskip{12pt}
\setlength\abovedisplayshortskip{0pt}
\setlength\belowdisplayshortskip{10pt}
\begin{align}
&\toktag{How}{QUE} \toktag{tall}{DEG} \toktag{is}{NOW} \toktag{the}{DEF}
\toktag{green\tokglue monster}{ART} \toktag{at}{REL} \toktag{Fenway}{GEO}
\toktag{?}{QUE}
\label{sem-tag:1}
\\
&\toktag{My}{HAS} \toktag{sister}{ROL} \toktag{went}{EPS} \toktag{to}{REL} \toktag{the}{DEF} \toktag{United\tokglue States}{GPE} \toktag{to}{SUB} \toktag{study}{EXS} \toktag{English}{CON} \toktag{.}{NIL}
\label{sem-tag:2}
\\
&\toktag{Any}{AND} \toktag{contribution}{CON} \toktag{was}{PST} \toktag{appreciated}{EXS} \toktag{but}{BUT} \toktag{we}{PRO} \toktag{have}{NOW} \toktag{n't}{NOT} \toktag{got}{EXT} \toktag{any}{DIS} \toktag{.}{NIL}
\label{sem-tag:3}
\\
&\toktag{He}{PRO} \toktag{himself}{EMP} \toktag{can}{POS} \toktag{earn}{EXS} \toktag{\$}{UOM} \toktag{100}{QUC} \toktag{a}{AND} \toktag{day}{UOM} \toktag{.}{NIL}
\label{sem-tag:4}
\end{align}

\comments{
\noindent\begin{minipage}{\textwidth}
\small
\toktag{How}{QUE} \toktag{tall}{DEG} \toktag{is}{NOW} \toktag{the}{DEF}
\toktag{green\tokglue monster}{ART} \toktag{at}{REL} \toktag{Fenway}{GEO}
\toktag{?}{QUE}
\toktag{My}{HAS} \toktag{sister}{ROL} \toktag{went}{EPS} \toktag{to}{REL} \toktag{the}{DEF} \toktag{United\tokglue States}{GPE} \toktag{to}{SUB} \toktag{study}{EXS} \toktag{English}{CON} \toktag{.}{NIL}
\toktag{Any}{AND} \toktag{contribution}{CON} \toktag{was}{PST} \toktag{appreciated}{EXS} \toktag{but}{BUT} \toktag{we}{PRO} \toktag{have}{NOW} \toktag{n't}{NOT} \toktag{got}{EXT} \toktag{any}{DIS} \toktag{.}{NIL}
\toktag{He}{PRO} \toktag{himself}{EMP} \toktag{can}{POS} \toktag{earn}{EXS} \toktag{\$}{UOM} \toktag{100}{QUC} \toktag{a}{AND} \toktag{day}{UOM} \toktag{.}{NIL}
\end{minipage}
}
\egroup

\section{Applications and Results}
\label{sec:pmb_app}

The idea of the universal semantic tagging was originally motivated by the goals of the PMB project \citep{Bos2014ISA,PMBshort:2017}:
(i) compositionally derive formal meaning representations for wide-coverage English text \citep{Bos2009GSCL}, and (ii) project the meaning representations to Dutch, German and Italian translations via word alignments \citep{evang2016}.  
These requirements challenge semantic competence and cross-lingual scalability of the universal semantic tagging.

A high quality large-scale semantic lexicon is crucial for the PMB as both projection and derivation of meaning representations starts from lexical items.
The semantic tagset plays a crucial role in development and organization of the lexicon.
In particular, in the PMB, Boxer \citep{Bos2008STEP2,boxer} interprets a sem-tag as a mapping from CCG categories (augmented with thematic roles) to a formal semantic schema which is further specified by a token-related predicate/constant symbol and thematic roles (if any).
The function behind the \smtg{EXS} tag is partially depicted in \eqref{eq:EXS}:

\bgroup
\setlength\abovedisplayskip{2pt}
\setlength\belowdisplayskip{12pt}
\setlength\abovedisplayshortskip{0pt}
\setlength\belowdisplayshortskip{10pt}
\begin{equation}
\scalebox{.95}{
$\mathtt{EXS} = \left \{ 
\begin{tabular}{r@{~~~}c@{~~~}l}
$S\bs_{R_1}\NP$ & $\mapsto$ &
$\lam{P \, r}  P \big( \lam{x} \scalebox{.75}{
\drs{$e$}{$\text{SYM}(e)$ ~ $R_1(e,x)$} 
} ; r(e) \big)$
\\
$(S\bs_{R_1}\NP)/_{R_2}\NP$ & $\mapsto$ &
$\lam{Q \, P \, r} P \Big( \lam{x} Q \big(\lam{y} \scalebox{.75}{
\drs{$e$}{$\text{SYM}(e)$ ~ $R_1(e,x)$ ~ $R_2(e,y)$} 
}; r(e) \big) \Big)$
\end{tabular}
\right \}$
}
\label{eq:EXS}
\end{equation}
\egroup

In order to 
collect large semantically annotated data via bootstrapping, we prepared initial silver and gold datasets for system training and testing respectively.
The datasets are part of the PMB, where the gold data is manually checked and consists of 2.4K English sentences  (14.6K tokens) while the silver data (457K tokens) consists of the PMB documents that are tagged by the neural semantic tagger of \cite{Bjervaetal:16} and have some manual corrections. 
For the data collection via bootstrapping, we initially employ the tri-gram based TnT tagger \citep{Brants:2000} rather than data-hungry neural models.
After training TnT on the silver data, it correctly tagged 86.89\% of tokens in the gold data: almost 5\% improvement over the most frequent tag per-word baseline (82.18\%).
This accuracy seems promising for bootstrapping application.%
\footnote{This result of the TnT tagger is not directly comparable to the result (83.6\%) of the neural semantic tagger reported by \cite{Bjervaetal:16} since the experiments differ in terms of training/test data and the semantic tagset.
}

Besides the application in the PMB, \cite{Bjervaetal:16} showed that using sem-tags as auxiliary information significantly improves English Universal Dependencies \pos{} tagging.
Given that Boxer and similar semantic parsing scenarios are commonly used \citep{mineshima-EtAl:2015:EMNLP,Beltagy:16,lewis-steedman:13}, semantic tagging will help those researches to shift to a cross-lingual level.
Additionally, multilingual semantic parsing approaches might also benefit from semantic tagging.
For example, sematic tags can help UD{\sc ep}L{\sc ambda}\cite{reddy_USP_2017} to decrease efforts of looking up lexical information for several words, e.g., quantifiers and negation markers.

\section{Conclusion}

We have proposed a novel NLP task that contributes to wide-coverage cross-lingual semantic parsing.
Tagging tokens with universal semantic tags represents an independent task that unifies and generalizes over semantic virtues of \pos{}-tagging and \namen{} recognition.
The expressive semantic tagset allows disambiguation of various semantic phenomena.
Besides their application in semantic parsing, already demonstrated in the PMB project, sem-tags can contribute to other NLP tasks, e.g. \pos{} tagging, or research lines rooted in compositional semantics. 

In contrast to POS tagsets \citep{Marcus:1993,uni_pos_tagset} augmented with morphological/universal features \citep{unimorph:16,UD_v1:16}, the semantic tagset is less expressive from a morphological perspective.
On the other hand, the tagset is leaner and models several semantic phenomena, e.g., roles (\smtg{ROL}), subsectives (\smtg{SST}), privatives (\smtg{PRI}), and degrees (\smtg{DEG}), that are beyond morphology.
Compared to the standard \namen{} classes \citep{CoNLL2003:ner}, the named entity class (\mttg{NAM}) of the tagset is broader.
The annotations of temporal expressions at TempEval \citep{SemEval-2013-tempeval} and MUC-7 \citep{MUC7def:98} differ from semantic tagging in terms of granularity: they annotate entire time expressions (e.g., {\em August of 2014}) while the semantic tagset opts for a more compositional analysis.

In future research, we plan to annotate more data with the help of human annotators, automatically tag large monolingual data via bootstrapping, further improve cross-lingual projection of sem-tags, and prepare an annotation guideline for semantic tagging. 
Elaboration of compositional semantics in the PMB might lead to an additional refinement of the semantic tagset.
For example, one can distinguish genders or animacy for better pronoun resolution or mark plurality information for better semantic analysis.

\section*{Acknowledgements}

This work has been supported by the NWO-VICI grant ``Lost in Translation – Found in Meaning'' (288-89-003). 
We also wish to thank the three anonymous reviewers for their helpful comments.

\bibliographystyle{chicago}
\bibliography{my_references}

\end{document}

%% file: tagset2.tex
\setlength{\aboverulesep}{.85pt}
\setlength{\belowrulesep}{.85pt}
\begin{tabular}[t]{R{2.5cm}@{} p{7cm}}
\addlinespace[.5pt]
\cmidrule(r){2-2} 
\multirow{1}{*}{\tagdeftab{\metatag{ANA}}{anaphoric}} 
&\semtag{PRO} \tagdef{anaphoric \& deictic pronouns}:
	\tagex{he, she, I, him}\\
&\semtag{DEF} \tagdef{definite}:
	\tagex{the, lo$^{\text{IT}}$, der$^{\text{DE}}$}\\
&\semtag{HAS} \tagdef{possessive pronoun}:
	\tagex{my, her}\\
&\semtag{REF} \tagdef{reflexive \& reciprocal pron.}:
	\tagex{herself, each\tokglue{}other}\\
&\semtag{EMP} \tagdef{emphasizing pronouns}:
	\tagex{himself}   
\\\cmidrule(r){2-2} 
\multirow{1}{*}{\tagdeftab{\metatag{ACT}}{speech\\[-.8mm]act}} 
&\semtag{GRE} \tagdef{greeting \& parting}:
	\tagex{hi, bye}\\
&\semtag{ITJ} \tagdef{interjections, exclamations}:
	\tagex{alas, ah}\\
&\semtag{HES} \tagdef{hesitation}:
	\tagex{err}\\
&\semtag{QUE} \tagdef{interrogative}:
	\tagex{who, which, ?}
\\\cmidrule(r){2-2}  
\multirow{1}{*}{\tagdeftab{\metatag{ATT}}{attribute}} 
&\semtag{QUC}\newtag \tagdef{concrete quantity}:
	\tagex{two, six\tokglue million, twice}\\
&\semtag{QUV}\newtag \tagdef{vague quantity}:
	\tagex{millions, many, enough}\\
&\semtag{COL}\newtag \tagdef{colour}:
	\tagex{red, crimson, light\tokglue blue, chestnut\tokglue brown}\\
&\semtag{IST} \tagdef{intersective}:
	\tagex{open, vegetarian, quickly}\\
&\semtag{SST} \tagdef{subsective}:
	\tagex{skillful \context{surgeon}, tall \context{kid}}\\
&\semtag{PRI} \tagdef{privative}:
	\tagex{former, fake}\\
&\semtag{DEG}\newtag \tagdef{degree}:
	\tagex{\context{2 meters} tall, \context{20 years} old}\\
&\semtag{INT} \tagdef{intensifier}:
	\tagex{very, much, too, rather}\\
&\semtag{REL} \tagdef{relation}:
	\tagex{in, on, 's, of, after}\\
&\semtag{SCO} \tagdef{score}:
	\tagex{3-0, \context{grade} A}
\\\cmidrule(r){2-2} 
\multirow{1}{*}{\tagdeftab{\metatag{COM}}{com-\\[-.8mm]parative}} 
&\semtag{EQU} \tagdef{equative}:
	\tagex{as \context{tall as John}, \context{whales} are \context{mammals}}\\
&\semtag{MOR} \tagdef{comparative positive}:
	\tagex{better, more}\\
&\semtag{LES} \tagdef{comparative negative}:
	\tagex{less, worse}\\
&\semtag{TOP} \tagdef{superlative positive}:
	\tagex{most, mostly}\\
&\semtag{BOT} \tagdef{superlative negative}:
	\tagex{worst, least}\\
&\semtag{ORD} \tagdef{ordinal}:
	\tagex{1st, 3rd, third}
\\\cmidrule(r){2-2} 
\multirow{1}{*}{\tagdeftab[r]{\metatag{UNE}}{unnamed\\[-.8mm]entity}}
&\semtag{CON} \tagdef{concept}:
	\tagex{dog, person}\\
&\semtag{ROL} \tagdef{role}:
	\tagex{student, brother, prof., victim}\\
&\semtag{GRP}\newtag \tagdef{group}:
	\tagex{\context{John} \{,\} \context{Mary} and \context{Sam gathered}, \context{a} group \context{of people}}
\\\cmidrule(r){2-2} 
\multirow{1}{*}{\tagdeftab{\metatag{DXS}}{deixis}}
&\semtag{DXP}\newtag \tagdef{place deixis}: 
	\tagex{here, this, above}\\
&\semtag{DXT}\newtag \tagdef{temporal deixis}:
	\tagex{just, later, tomorrow}\\
&\semtag{DXD}\newtag \tagdef{discourse deixis}:
	\tagex{latter, former, above}
\\\cmidrule(r){2-2} 
\multirow{1}{*}{\tagdeftab{\metatag{LOG}}{logical}} 
&\semtag{ALT} \tagdef{alternative \& repetitions}:
	\tagex{another, different, again}\\
&\semtag{XCL} \tagdef{exclusive}:
	\tagex{only, just}\\
&\semtag{NIL} \tagdef{empty semantics}:
	\tagex{\{.\}, to, of}\\
&\semtag{DIS} \tagdef{disjunction \& exist. quantif.}:
	\tagex{a, some, any, or}\\
&\semtag{IMP} \tagdef{implication}:
	\tagex{if, when, unless}\\
&\semtag{AND} \tagdef{conjunction \& univ. quantif.}:
	\tagex{every, and, who, any}
\\\cmidrule(r){2-2}
\end{tabular}
\hspace{-4mm}
\setlength{\aboverulesep}{1pt}
\setlength{\belowrulesep}{1pt}
\begin{tabular}[t]{L{6.8cm}@{} p{2.5cm}}
\cmidrule(l){1-1} 
\semtag{NOT} \tagdef{negation}:
	\tagex{not, no, neither, without}
&\multirow{1}{*}{\tagdeftab[l]{\metatag{MOD}}{modality}}\\
\semtag{NEC} \tagdef{necessity}:
	\tagex{must, should, have \context{to}}&\\
\semtag{POS} \tagdef{possibility}:
	\tagex{might, could, perhaps, alleged, can}&\\
\addlinespace[1pt]
\cmidrule(l){1-1} 
\addlinespace[1pt]        
\semtag{SUB} \tagdef{subordinate relations}:
	\tagex{that, while, because}
&\multirow{1}{*}{\tagdeftab[l]{\metatag{DSC}}{discourse}}\\
\semtag{COO} \tagdef{coordinate relations}:
	\tagex{so, \{,\}, \{;\}, and}&\\
\semtag{APP} \tagdef{appositional relations}:
	\tagex{\{,\}, which, \{(\}, \{---\}}&\\
\semtag{BUT} \tagdef{contrast}:
	\tagex{but, yet}&\\
\addlinespace[1pt]
\cmidrule(l){1-1} 
\addlinespace[1pt]
\semtag{PER} \tagdef{person}:
	\tagex{Axl\tokglue Rose, Sherlock\tokglue Holmes}
&\multirow{1}{*}{\tagdeftab[l]{\metatag{NAM}}{named\\[-.8mm]entity}}\\
\semtag{GPE} \tagdef{geo-political entity}:
	\tagex{Paris, Japan}&\\
\semtag{GPO}\newtag \tagdef{geo-political origin}:
	\tagex{Parisian, French}&\\
\semtag{GEO} \tagdef{geographical location}:
	\tagex{Alps, Nile}&\\
\semtag{ORG} \tagdef{organization}:
	\tagex{IKEA, EU}&\\
\semtag{ART} \tagdef{artifact}:
	\tagex{iOS\tokglue 7}&\\
\semtag{HAP} \tagdef{happening}:
	\tagex{Eurovision\tokglue 2017}&\\
\semtag{UOM} \tagdef{unit of measurement}:
	\tagex{meter, \$, \%, degree\tokglue Celsius\kern-5mm}&\\
\semtag{CTC}\newtag \tagdef{contact information}:
	\tagex{112, info@mail.com}&\\
\semtag{URL} \tagdef{URL}: 
	\tagex{\scriptsize\url{http://pmb.let.rug.nl}}&\\
\semtag{LIT}\newtag \tagdef{literal use of names}:
	\tagex{\context{his name is} John}&\\
\semtag{NTH}\newtag \tagdef{other names}:
	\tagex{\context{table} 1a, \context{equation} (1)}    
\\\cmidrule(l){1-1} 
\semtag{EXS} \tagdef{untensed simple}:
	\tagex{\context{to} walk, \context{is} eaten, destruction}
&\multirow{1}{*}{\tagdeftab[l]{\metatag{EVE}}{\tagdef{events}}}\\    
\semtag{ENS} \tagdef{present simple}:
	\tagex{\context{we} walk, \context{he} walks}&\\
\semtag{EPS} \tagdef{past simple}:
	\tagex{ate, went}&\\
\semtag{EXG} \tagdef{untensed progressive}:
	\tagex{\context{is} running}&\\
\semtag{EXT} \tagdef{untensed perfect}:
	\tagex{\context{has} eaten}&\\
\addlinespace[2pt]
\cmidrule(l){1-1} 
\addlinespace[2pt]
\semtag{NOW} \tagdef{present tense}:
	\tagex{is \context{skiing}, do \context{ski}, has \context{skied}, now}
&\multirow{1}{*}{\tagdeftab[l]{\metatag{TNS}}{\tagdef{tense \&\\[-.8mm]aspect}}}\\
\semtag{PST} \tagdef{past tense}:
	\tagex{was \context{baked}, had \context{gone}, did \context{go}}&\\
\semtag{FUT} \tagdef{future tense}:
	\tagex{will, shall}&\\
\semtag{PRG}\newtag \tagdef{progressive}:
	\tagex{\context{has been} being \context{treated}, aan\tokglue het$^\text{NL}$}&\\
\semtag{PFT}\newtag \tagdef{perfect}:
	\tagex{\context{has} been \context{going/done}}&\\
\addlinespace[2pt]
\cmidrule(l){1-1} 
\addlinespace[2pt] 
\semtag{DAT}\newtag \tagdef{full date}: 
	\tagex{27.04.2017, 27/04/17}
&\multirow{1}{*}{\tagdeftab[l]{\metatag{TIM}}{\tagdef{temporal\\[-.8mm]entity}}}\\    
\semtag{DOM} \tagdef{day of month}:
	\tagex{27th \context{December}}&\\
\semtag{YOC} \tagdef{year of century}:
	\tagex{2017}&\\
\semtag{DOW} \tagdef{day of week}:
	\tagex{Thursday}&\\
\semtag{MOY} \tagdef{month of year}:
	\tagex{April}&\\
\semtag{DEC} \tagdef{decade}:
	\tagex{80s, 1990s}&\\
\semtag{CLO} \tagdef{clocktime}:
	\tagex{8:45\tokglue pm, 10\tokglue o'clock, noon}&
\\\cmidrule(l){1-1} 
\end{tabular}